\documentclass[runningheads]{llncs}
\usepackage[T1]{fontenc}
\usepackage{graphicx,verbatim}
\usepackage{bm}
\usepackage{amssymb,amsmath}
\usepackage{tabularx,booktabs}
\usepackage{makecell}

\usepackage{hyperref}
\usepackage{color}

\begin{document}
\title{Lesion-Aware Post-Training of Latent Diffusion Models for Synthesizing Diffusion MRI from CT Perfusion}
\author{%
Junhyeok Lee\inst{1\star}% index{Lee, Junhyeok}
\and
Hyunwoong Kim\inst{2\star}% index{Kim, Hyunwoong}
\and
Hyungjin Chung\inst{3}% index{Chung, Hyungjin}
\and
Heeseong Eom\inst{1}% index{Eom, Heeseong}
\and \\
Joon Jang\inst{4}% index{Jang, Joon}
\and
Chul-Ho Sohn\inst{2}% index{Sohn, Chul-Ho}
\and
Kyu Sung Choi\inst{2\dag}% index{Choi, Kyu Sung}
}
\authorrunning{J. Lee et al.}
\titlerunning{Lesion-Aware Post-Training of Latent Diffusion Models}
%\orcidID{0000-1111-2222-3333}
% \url{http://www.springer.com/gp/computer-science/lncs} 
\institute{College of Medicine, Seoul National University, Seoul, Republic of Korea \and
Department of Radiology, Seoul National University Hospital,\\Seoul, Republic of Korea \and
Department of Bio and Brain Engineering, Korea Advanced Institute of \\Science and Technology, Daejeon, Republic of Korea \and
Department of Biomedical Sciences, Seoul National University,\\Seoul, Republic of Korea \\
\email{\{ent1127@snu.ac.kr\}}}
\maketitle   
\renewcommand{\thefootnote}{\fnsymbol{footnote}}
\footnotetext[1]{These authors contributed equally to this paper.}

% \end{comment}

\begin{comment}
\author{Anonymized Authors}  %% Added for anonymized MICCAI 2025 submission
\authorrunning{Anonymized Author et al.}
\titlerunning{Lesion-Aware Post-Training of Latent Diffusion Models}
\institute{Anonymized Affiliations \\
    \email{email@anonymized.com}}
\maketitle  
\end{comment}

\begin{abstract}
Image-to-Image translation models can help mitigate various challenges inherent to medical image acquisition. Latent diffusion models (LDMs) leverage efficient learning in compressed latent space and constitute the core of state-of-the-art generative image models. However, this efficiency comes with a trade-off, potentially compromising crucial pixel-level detail essential for high-fidelity medical images. This limitation becomes particularly critical when generating clinically significant structures, such as lesions, which often occupy only a small portion of the image. Failure to accurately reconstruct these regions can severely impact diagnostic reliability and clinical decision-making. To overcome this limitation, we propose a novel post-training framework for LDMs in medical image-to-image translation by incorporating lesion-aware medical pixel space objectives. This approach is essential, as it not only enhances overall image quality but also improves the precision of lesion delineation. We evaluate our framework on brain CT-to-MRI translation in acute ischemic stroke patients, where early and accurate diagnosis is critical for optimal treatment selection and improved patient outcomes. While diffusion MRI is the gold standard for stroke diagnosis, its clinical utility is often constrained by high costs and low accessibility. Using a dataset of 817 patients, we demonstrate that our framework improves overall image quality and enhances lesion delineation when synthesizing DWI and ADC images from CT perfusion scans, outperforming existing image-to-image translation models. Furthermore, our post-training strategy is easily adaptable to pre-trained LDMs and exhibits substantial potential for broader applications across diverse medical image translation tasks.

% Authors must provide keywords and are not allowed to remove this Keyword section.

\end{abstract}
\section{Introduction}
Medical imaging plays a crucial role in modern medicine, providing spatially resolved information of organs and tissues. Various imaging modalities offer unique clinical insights with distinct advantages and limitations based on their underlying physical principles. In the context of acute stroke management where "time is brain", rapid imaging for diagnosis is crucial as timely intervention directly impacts patient outcomes \cite{doi:10.1161/01.STR.0000196957.55928.ab,https://doi.org/10.1111/jon.12693}. Computed tomography (CT) is frequently utilized due to its widespread availability, short acquisition times, and low cost. Although CT can detect early signs of acute ischemic stroke, these indicators are often subtle or absent within the initial hours following stroke onset, leading to suboptimal sensitivity and inter-rater agreement \cite{Bryan611}. In contrast, diffusion-weighted imaging (DWI) on magnetic resonance imaging (MRI) offers superior sensitivity for detecting acute ischemic stroke and distinguishing stroke mimics \cite{Stroke_CT_DWI,Chalela2007}. However, MRI has several limitations compared to CT, including higher costs, restricted accessibility, longer scan durations, and challenges related to patient intolerance or contraindications. 

With the recent explosion of image generative models, there has been wide interest in medical image-to-image translation model development to help overcome challenges in medical image acquisition \cite{ARMANIOUS2020101684,10484171}. In recent years, diffusion models have surpassed generative adversarial networks (GANs) as state-of-the-art image generation models \cite{dhariwal2021diffusion}. Latent diffusion models (LDMs) have emerged as a particularly efficient approach, operating within a compressed latent space to improve both computational efficiency and generative performance \cite{rombach2022highresolutionimagesynthesislatent}. Moreover, LDMs demonstrated promising results in various medical imaging applications, including image synthesis \cite{Müller-Franzes2023}, super-resolution \cite{li2024rethinkingdiffusionmodelmulticontrast}, and image-to-image translation \cite{10484171}. However, LDMs learn the diffusion process only in the latent space and often freeze the decoder, potentially overlooking high-frequency image details. Existing methods have limitations in addressing this challenge effectively. Only recently, few studies have explored post-training techniques for LDMs incorporating image space objectives for sharper and more realistic image generation \cite{berrada2025boostinglatentdiffusionperceptual,zhang2024pixelspaceposttraininglatentdiffusion}. In medical imaging, this challenge intensifies when generating clinically significant structures, such as lesions, which often show low spatial occupancy. Deficiencies in this reconstruction can substantially degrade diagnostic reliability and subsequent clinical decision-making.

\begin{figure}
    \includegraphics[width=\textwidth]{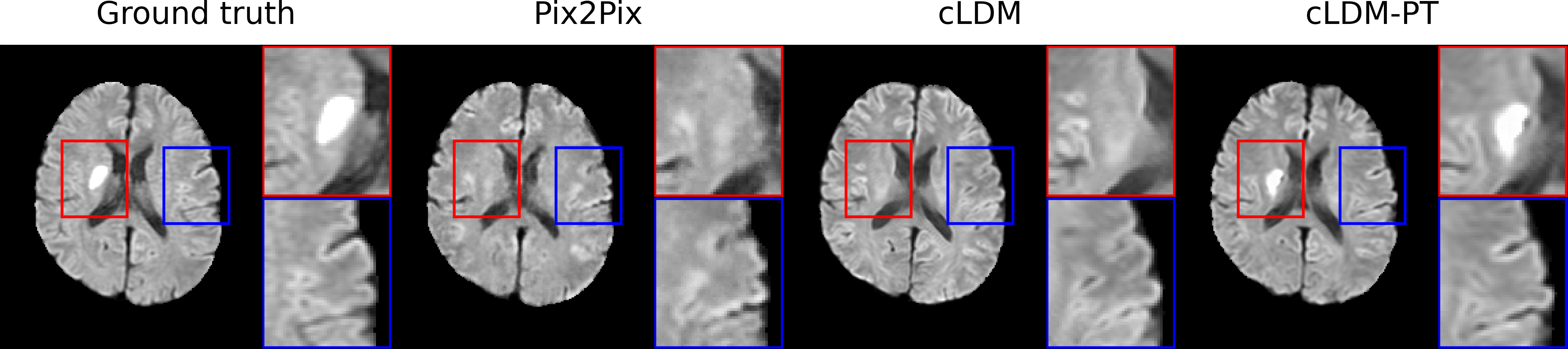}
    \caption{Previous models fail to accurately depict ischemic stroke lesion in diffusion MRI synthesized from CT perfusion. Our model (right) shows higher lesion conspicuity (red) and enhanced image fidelity, highlighted with grey-white matter differentiation (blue).} \label{fig1}
\end{figure}

In this study, we propose a novel post-training framework for LDMs in medical image-to-image translation with lesion-aware medical image space objectives. Our method incorporates medical image space loss to generate more realistic medical images. In addition to the pixel loss for overall image quality, we introduce a task-specific objective for ischemic lesion areas to enhance the accuracy of lesion delineation. Evaluation on a diffusion MRI-CT perfusion paired dataset from 817 acute ischemic stroke patients demonstrate that our LDM post-training framework outperforms state-of-the-art models in both qualitative and quantitative evaluations. Moreover, we apply our framework to other diffusion models that utilizes the latent space, demonstrating its adaptability and potential for broad utility in diverse medical image generative tasks.

\section{Method}

\subsection{Base Conditional Latent Diffusion Model}
Given an axial slice of target MRI images $\bm{x} \in \mathbb{R}^{H \times W \times m}$ with $m$ modalities concatenated into channels, our base model is a conditional LDM with the VQGAN \cite{esser2021tamingtransformershighresolutionimage} framework where the encoder $\mathcal{E}$ encodes the images into latent representations $\bm{z} = \mathcal{E}(\bm{x})$. The decoder $\mathcal{D}$ learns to reconstruct latent representations back into MRI images as $\bm{x} = \mathcal{D}(\bm{z})$.
The corresponding axial slice of the CTP image $\bm{c} \in \mathbb{R}^{H \times W \times n}$ with $n$ time points is used as the input condition for the CTP-to-MRI translation diffusion process. To align the CTP image $\bm{c}$ with the latent representation $\bm{z}$ of MRI images, we follow the standard procedure for semantic synthesis with LDMs \cite{rombach2022highresolutionimagesynthesislatent}. A bilinear interpolator combined with 1$\times$1 convolutions is used as a spatial rescaler $\mathcal{F}$ to downsample $\bm{c}$ into $\tilde{\bm{c}}=\mathcal{F}(\bm{c})$, then $\tilde{\bm{c}}$ is fed into the diffusion process by channel-wise concatenation.

The conditional LDM learns the latent distribution of the MRI images $p_\theta(\bm{z})$ by learning denoising conditional autoencoders $\bm{\epsilon}_{\theta}(\bm{z}_t, t, \tilde{\bm{c}})$ from the sequence of noisy images $\{\bm{z}_0,...,\bm{z}_T\}$. The forward process that diffuses the latent input $\bm{z}_0=\bm{z}$ with pre-defined Gaussian noise schedules $\{\beta_1,...,\beta_T\}$ is a Markov process formulated as:
\begin{equation}
    q(\bm{z}_t|\bm{z}_{t-1}) = \mathcal{N}(\bm{z}_t; \sqrt{1-\beta_t}\bm{z}_{t-1}, \beta_t\mathbf{I}),
\end{equation}
which allows sampling during training via:
\begin{equation}
    q(\bm{z}_t|\bm{z}_0) = \mathcal{N}(\bm{z}_t; \sqrt{\bar{\alpha}_t}\bm{z}_0, (1-\bar{\alpha}_t)\mathbf{I}),
\end{equation}
where $\alpha_t = 1-\beta_t$ and $\bar{\alpha}_t = \prod_{i=1}^t \alpha_i$. The reverse process that denoises noisy images is formulated with a time-conditional UNet \cite{ronneberger2015unetconvolutionalnetworksbiomedical} with parameters $\theta$ as:
\begin{equation}
    p(\bm{z}_{t-1}|\bm{z}_t,\tilde{\bm{c}}) = \mathcal{N}(\bm{z}_{t-1}; \bm{\mu}_\theta(\bm{z}_t,t,\tilde{\bm{c}}), \bm{\Sigma}_\theta(\bm{z}_t,t,\tilde{\bm{c}})).
\end{equation}
After parameterizing $\bm{\mu}_\theta(\bm{z}_t,t,\tilde{\bm{c}}) = \frac{1}{\sqrt{\alpha_t}}(\bm{z}_t-\frac{1-\alpha_t}{\sqrt{1-\bar{a}_t}}\bm{\epsilon}_\theta(\bm{z}_t,t,\tilde{\bm{c}}))$ and simplifying $\bm{\Sigma}_\theta(\bm{z}_t,t,\tilde{\bm{c}}) = \sigma_t^2\mathbf{I}$, the latent objective for the conditional LDM trained to predict $\bm{\epsilon}$ is given as:
\begin{equation}
    \mathcal{L}_{latent} = \mathbb{E}_{\mathcal{E}(\bm{x}), \bm{\epsilon},t}[\|\bm{\epsilon}-\bm{\epsilon}_\theta(\bm{z}_t, t, \tilde{\bm{c}})\|_2^2].
\end{equation}
Detailed illustration of the model is shown in Figure \ref{fig1}.
% \subsection{Brownian Bridge Diffusion Model}
% BBDM is a variant of the latent diffusion model specialized for image-to-image translation between two domains using the Brownian Bridge diffusion process \cite{li2023bbdmimagetoimagetranslationbrownian}. 

\begin{figure}[t]
    \includegraphics[width=\textwidth]{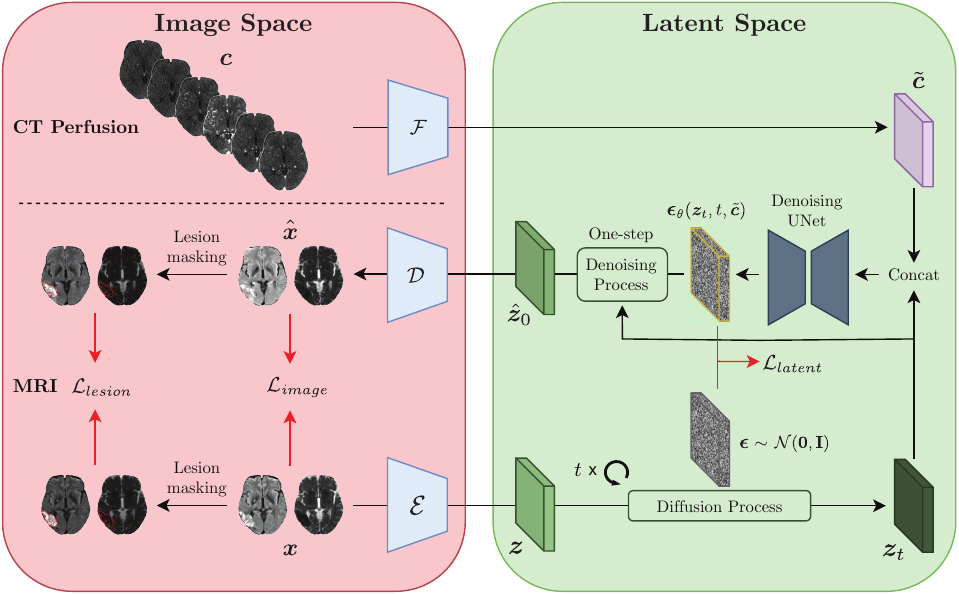}
    \caption{\textbf{Overview of our post-training framework.} During post-training, in addition to the latent objective for the conditional LDM $\mathcal{L}_{latent}$, we introduce medical image space objectives $\mathcal{L}_{image}$, $\mathcal{L}_{lesion}$ to enhance overall image quality and lesion conspicuity.}\label{fig2}
\end{figure}

\subsection{Lesion-Aware Image Space Objectives for LDM Post-Training}
We introduce medical image space objectives for LDM post-training to improve overall image quality and allow ischemic lesion-aware diffusion MRI generation. Recent studies show post-training LDMs with image space objectives can lead to sharper and more realistic natural images \cite{berrada2025boostinglatentdiffusionperceptual,zhang2024pixelspaceposttraininglatentdiffusion}. For image generation in domains where high image detail is essential such as medical imaging \cite{ARMANIOUS2020101684}, remote sensing \cite{Sebaq2024}, or face generation \cite{10204758}, post-training LDMs with task-specific image space objectives can be particularly beneficial. 

Given a MRI image $\bm{x}$ and its latent representation $\bm{z} = \mathcal{E}(\bm{x})$, the noisy version of $\bm{z}$ is sampled as $\bm{z}_t = \sqrt{\bar{\alpha}_t}\bm{z} + \sqrt{1-\bar{\alpha}_t}\bm{\epsilon}$ during training. We then project the estimated noise free latent $\hat{\bm{z}}_0$ given as:
\begin{equation}
    \hat{\bm{z}}_0 = \frac{\bm{z}_t - \sqrt{1-\bar{\alpha}_t}\bm{\epsilon}_\theta(\bm{z}_t, t, \tilde{\bm{c}})}{\sqrt{\bar{\alpha}_t}},
\end{equation} 
back to the image space using the decoder $\mathcal{D}$ to get $\hat{\bm{x}} = \mathcal{D}(\hat{\bm{z}}_0)$. This formulation allows fast and efficient one-step inference of $\hat{\bm{z}}_0$ and thus $\hat{\bm{x}}$ during training instead of the iterative denoising steps of the standard inference process \cite{berrada2025boostinglatentdiffusionperceptual,ho2020denoisingdiffusionprobabilisticmodels}. The image space objective is then defined as:

\begin{equation}
    \mathcal{L}_{image} = \mathbb{E}_{\mathcal{E}(\bm{x}), \bm{\epsilon},t}[\|\bm{x} - \mathcal{D}(\hat{\bm{z}}_0)\|_2^2].
\end{equation}
One of the main challenges of CTP-to-MRI translation is training the model to accurately generate ischemic lesions that constitute only a small portion of the voxels in the dataset. We believe standard objectives covering the entire image or latent representations are insufficient to train the model to precisely generate lesions because the gradients to guide the model to generate lesions is diluted with signals from other brain parenchyma or even less important background voxels.
To boost accurate ischemic lesion generation, we designed a new image space objective focusing only on ischemic lesion regions. The lesion-aware objective is formulated as:
\begin{equation}
    \mathcal{L}_{lesion} = \mathbb{E}_{\mathcal{E}(\bm{x}), \bm{\epsilon},t}[\|M_{lesion}(\bm{x}) \odot (\bm{x} - \mathcal{D}(\hat{\bm{z}}_0))\|_2^2],
\end{equation}
where $M_{lesion}(\bm{x}) \in \mathbb{R}^{H \times W}$ is a binary mask for the ischemic lesion. The final objective function combining the latent and image space losses with hyper-parameters $\lambda_{image}, \lambda_{lesion}$ is given as:
\begin{equation}
    \mathcal{L}_{total} = \mathcal{L}_{latent} + \lambda_{image}  \cdot \mathcal{L}_{image} + \lambda_{lesion} \cdot \mathcal{L}_{lesion}.
\end{equation}
\section{Experiments}
\subsection{Experimental Setup}
\subsubsection{Dataset}
All images used in this study were collected from the Seoul National University Hospital (SNUH) with approval from its Institutional Review Board. MRI scans including diffusion-weighted imaging (DWI) were acquired using 3.0T scanners with voxel size of 0.9375×0.9375 mm to 1×1 mm (in-plane) and 4–5 mm (axial slice thickness). The images were skull-stripped then resampled to a uniform voxel size of 1×1×5 mm\textsuperscript{3}. 

Apparent diffusion coefficient (ADC) maps were derived from the DWI image ($b$=1000), and was used for ischemic lesion segmentation by medical experts to create ground truth lesion masks. CTP scans were acquired using Aquilion 64 CT scanner (TOSHIBA) with voxel size of 0.47×0.47×1 mm\textsuperscript{3} and spanning 15 time points. To align CTP images with MRI images, CTP images were skull-stripped then registered into the DWI space using ANTs \cite{tustison_antsx_2021}. 

The final dataset comprised of paired DWI, ADC, and CTP images with ischemic lesion masks from 817 patients. The dataset were randomly divided into training (571 patients; 14083 axial slices), validation (81 patients; 1948 axial slices), and test (165 patients; 4015 axial slices) sets. Across the training, validation, and test sets, slices containing lesions constituted 10.9\% (1542/14083), 11.3\% (220/1948), and 11.0\% (441/4015), respectively, with mean (±SD) lesion volumes of 15.17±41.17 ml, 12.46±32.94 ml, and 14.66±36.24 ml.

\subsubsection{Implementation Details}
Implementation of our model was based on the LDM \cite{rombach2022highresolutionimagesynthesislatent} framework with VQGAN \cite{esser2021tamingtransformershighresolutionimage} that compresses the image space into the latent space by a factor of 4 in both coronal and sagittal directions. The model was trained with the AdamW \cite{loshchilov2018decoupled} optimizer with base learning rate of $2\times10^{-6}$. The model was trained with $T=1000$ diffusion steps, with 200 DDIM sampling steps used during inference. For LDM post-training, $\lambda_{image} = 0.01, 0.05,0.1$ were tested and several $\lambda_{lesion}$ values were selected accordingly. We selected $\lambda_{image} = 0.01, \lambda_{lesion}=0.02$ as the hyper-parameters of our best model. All experiments were conducted using the NVIDIA A6000 GPU with a batch size of 48. Our code is publicly available at \url{https://github.com/snuh-rad-aicon/Diffusion-LAPT}.
% Full code implementations used in this study will be released upon acceptance.

\begin{figure}[t]
    \centering
    \includegraphics[width=1\textwidth]{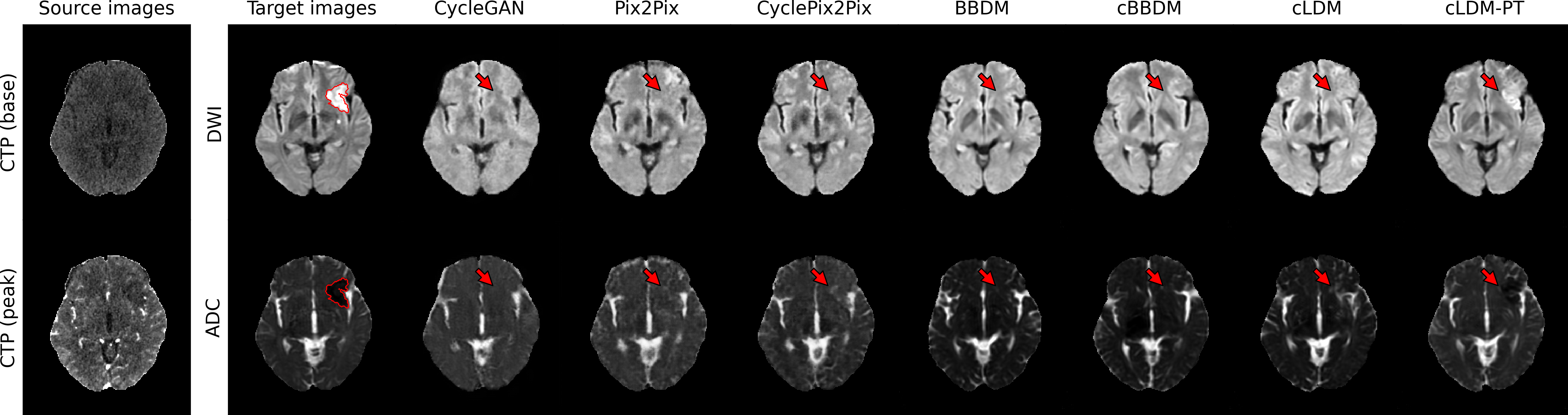}

    \includegraphics[width=1\textwidth]{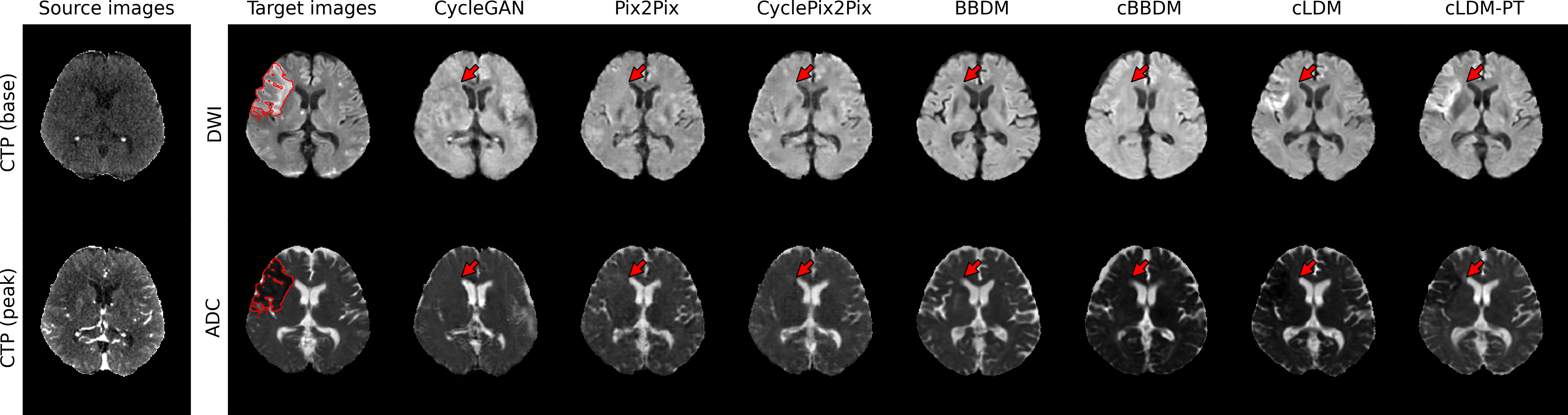}

    \includegraphics[width=1\textwidth]{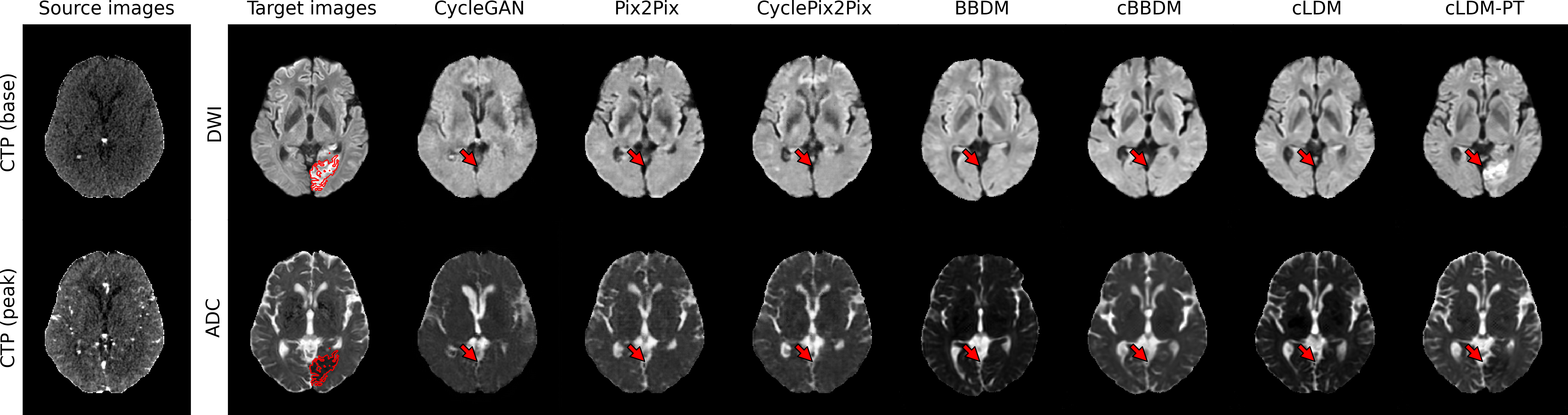}
    
    \caption{\textbf{Visualization of synthesized diffusion MRI images from CTP images in acute ischemic stroke patients.} Our model with post-training (cLDM-PT) excels in lesion delineation (red arrows), accurately depicting ischemic stroke lesions with restricted diffusion (red contour) based on hypo-perfused regions in source CTP images. (Top) A case with infarct core in the left inferior frontal area. (Middle) A case of acute ischemic stroke by large vessel occlusion in the right middle cerebral artery. (Bottom) A case of acute infarction in the left occipital lobe.
} 
    \label{fig3}
\end{figure}
% due to M2 superior division occlusion

\subsubsection{Evaluation Metrics}
To evaluate diffusion MRI synthesis, we used both distortion and perception  measures to compare ground truth ${\bm{x}}$ with synthesized image $\hat{\bm{x}}$. The mean absolute error (MAE) measures accuracy of image reconstruction, and to evaluate lesion delineation we additionally defined lesion MAE as: $\text{Lesion MAE} = (\sum M_{lesion}(\bm{x})\odot|\bm{x}-\hat{\bm{x}}|)/\sum M_{lesion}(\bm{x})$. Peak signal-to-noise ratio (PSNR) assesses reconstruction quality. The multi-scale structural similarity index measure (MS-SSIM) measures the similarity between images at multiple resolutions. The perceptual quality of the synthesized MRI images was measured using the Fréchet Inception Distance (FID) \cite{heusel2018ganstrainedtimescaleupdate}.

\subsubsection{Baselines}
We compared the performance of our model with state-of-the-art models, including models based on generative adversarial networks (Pix2Pix \cite{isola2018imagetoimagetranslationconditionaladversarial}, CycleGAN \cite{zhu2020unpairedimagetoimagetranslationusing}, PairedCycleGAN \cite{8578110}) and latent diffusion (conditional LDM \cite{rombach2022highresolutionimagesynthesislatent}, BBDM \cite{li2023bbdmimagetoimagetranslationbrownian}, conditional BBDM \cite{kim2024conditionalbrownianbridgediffusion}).

\subsection{Experimental Results}

\begin{table}[t]
    \caption{Quantitative results of DWI and ADC synthesis from CTP.}\label{tab:performance}
    \centering
    {\fontsize{8}{9.6}\selectfont
    \renewcommand{\arraystretch}{1} 
    \begin{tabularx}{\textwidth}{X >{\centering\arraybackslash}p{0.865cm} >{\centering\arraybackslash}p{0.865cm} >{\centering\arraybackslash}p{0.865cm} >{\centering\arraybackslash}p{0.865cm} >{\centering\arraybackslash}p{0.865cm} >{\centering\arraybackslash}p{0.865cm} >{\centering\arraybackslash}p{0.865cm} >{\centering\arraybackslash}p{0.865cm} >{\centering\arraybackslash}p{0.865cm} >{\centering\arraybackslash}p{0.865cm}}
        \toprule
        & \multicolumn{5}{c}{\textbf{DWI}} & \multicolumn{5}{c}{\textbf{ADC}} \\
        \cmidrule(lr){2-6} \cmidrule(lr){7-11}
        \textbf{Model} & MAE$\downarrow$ & \makecell{Lesion\\MAE$\downarrow$} & PSNR$\uparrow$ & \makecell{MS-\\SSIM$\uparrow$} & FID$\downarrow$ & MAE$\downarrow$ & \makecell{Lesion\\MAE$\downarrow$} & PSNR$\uparrow$ & \makecell{MS-\\SSIM$\uparrow$} & FID$\downarrow$ \\
        \midrule
        CycleGAN & 0.143 & 0.235 & 20.04 & 0.803 & 40.62 & 0.321 & 0.151 & 12.20 & 0.602 & \underline{65.95} \\
        Pix2Pix & 0.083 & \underline{0.215} & 26.25 & 0.859 & 46.90 & 0.073 & 0.127 & 26.52 & 0.872 & 70.64 \\
        PairedCycleGAN & 0.123 & 0.223 & 21.86 & \underline{0.865} & 47.96 & 0.081 & \underline{0.124} & 25.55 & \textbf{0.883} & 70.34 \\
        BBDM & 0.104 & 0.225 & 24.19 & 0.787 & 31.84 & 0.103 & 0.136 & 23.10 & 0.764 & 66.81 \\
        cBBDM & 0.093 & 0.226 & 25.21 & 0.800 & 32.32 & 0.098 & 0.139 & 23.63 & 0.777 & 66.98 \\
        cLDM & \underline{0.073} & 0.222 & \textbf{27.94} & 0.855 & \underline{30.82} & \underline{0.072} & 0.125 & \underline{27.08} & 0.866 & 66.71 \\
        \textbf{cLDM-PT(ours)} & \textbf{0.072} & \textbf{0.199} & \underline{27.78} & \textbf{0.867} & \textbf{29.95} & \textbf{0.052} & \textbf{0.105} & \textbf{31.49} & \underline{0.876} & \textbf{61.91} \\
        \bottomrule
    \end{tabularx}
    }
\end{table}

\subsubsection{Quantitative Results}
We observe a clear improvement from conventional methods in our refined latent diffusion model, cLDM-PT, which was optimized via our proposed post-training framework (Table \ref{tab:performance}). For both DWI and ADC generation, our cLDM-PT model achieved the lowest MAE, highest MS-SSIM, and lowest FID, indicating marked improvements in accuracy, clarity, and structural fidelity. Furthermore, our model achieved the lowest lesion MAE of 0.199 for DWI and 0.105 for ADC images, which underlines its enhanced capability for precise lesion delineation. Overall, our model showed 14.5\% deduction in image MAE and 12.4\% deduction in lesion MAE after post-training.

\subsubsection{Qualitative Evaluation}
Figure \ref{fig3} visualizes synthesized DWI and ADC from CTP of acute ischemic stroke patients with lesions in various brain regions. Due to low signal-to-noise ratio of CTP, it is difficult to accurately estimate ischemic core volumes. Small infarcts such as lacunar infarcts are also poorly visualized in CTP. These factors make it challenging for generative models to accurately reconstruct ischemic lesions in synthesized MRI. While the diffusion model series generates more realistic images compared to GAN-based models, they encounter difficulty in lesion delineation. Our model, cLDM-PT, excels in lesion delineation and demonstrates exceptional ability to generate accurate and detailed images.

\subsubsection{Impact of Loss Weights}
We test the effect of varying the weights of image space objectives  during post-training, defined as hyper-parameters $\lambda_{image}$ and $\lambda_{lesion}$ (Table \ref{tab:loss_performance}). Increasing $\lambda_{image}$ improves overall image accuracy and structural consistency, but this effect diminishes as $\lambda_{image}$ is further increased. As we increase $\lambda_{lesion}$, lesion MAE decreases sharply before reaching a plateau. Further increasing $\lambda_{lesion}$ results in overall distortion of synthesized images. We observe that small weights for image space objectives is sufficient to achieve the best balance between global image quality and lesion conspicuity.

\subsubsection{Generalization to Other Latent Diffusion Models}
To showcase the broad applicability of our framework, we apply it to cBBDM which learns stochastic Brownian bridge process in the latent space. Post-training cBBDM by our framework with $\lambda_{image}=0.01$ and $\lambda_{lesion}=0.02$ led to improvements across all metrics, indicating better image quality and lesion delineation (Table \ref{tab:cbbdm_performance}).

\begin{table}[t]
    \caption{Effect of image space loss weights on DWI and ADC synthesis from CTP.}
    \label{tab:loss_performance}
    \centering
    {\fontsize{8.}{9.6}\selectfont
    \renewcommand{\arraystretch}{1} 
    \begin{tabularx}{\textwidth}{>{\centering\arraybackslash}X >{\centering\arraybackslash}X >{\centering\arraybackslash}p{0.865cm} >{\centering\arraybackslash}p{0.865cm} >{\centering\arraybackslash}p{0.865cm} >{\centering\arraybackslash}p{0.865cm} >{\centering\arraybackslash}p{0.865cm} >{\centering\arraybackslash}p{0.865cm} >{\centering\arraybackslash}p{0.865cm} >{\centering\arraybackslash}p{0.865cm} >{\centering\arraybackslash}p{0.865cm} >{\centering\arraybackslash}p{0.865cm}}  
        \toprule
        \multicolumn{2}{c}{\textbf{Loss Weights}} & \multicolumn{5}{c}{\textbf{DWI}} & \multicolumn{5}{c}{\textbf{ADC}} \\
        \cmidrule(lr){1-2} \cmidrule(lr){3-7} \cmidrule(lr){8-12}
        $\lambda_{image}$ & $\lambda_{lesion}$ & MAE$\downarrow$ & \makecell{Lesion\\MAE$\downarrow$} & PSNR$\uparrow$ & \makecell{MS-\\SSIM$\uparrow$} & FID$\downarrow$ & MAE$\downarrow$ & \makecell{Lesion\\MAE$\downarrow$} & PSNR$\uparrow$ & \makecell{MS-\\SSIM$\uparrow$} & FID$\downarrow$ \\
        \midrule
        0 & 0   & \underline{0.073} & 0.222 & \textbf{27.94} & 0.855 & 30.82 & 0.072 & 0.125 & 27.08 & 0.866 & 66.71 \\
        \midrule
        0.01 & 0   & 0.074 & 0.202 & 27.48 & \textbf{0.867} & 31.08 & \underline{0.054} & 0.109 & \underline{30.89} & 0.876 & \underline{61.84} \\
        0.01 & 0.01 & 0.078 & 0.198 & 26.70 & \textbf{0.867} & 30.50 & \underline{0.054} & 0.106 & 30.76 & \underline{0.877} & 64.00 \\
        \textbf{0.01} & \textbf{0.02} & \textbf{0.072} & 0.199 & 27.78 & \textbf{0.867} & \textbf{29.95} & \textbf{0.052} & 0.105 & \textbf{31.49} & 0.876 & 61.91 \\
        0.01 & 0.05 & 0.077 & \underline{0.196} & 27.15 & \underline{0.865} & 30.74 & 0.058 & 0.093 & 29.33 & \textbf{0.878} & \textbf{61.52} \\
        \midrule
        0.05 & 0  & 0.077 & 0.212 & 27.05 & 0.863 & 30.66 & 0.058 & 0.116 & 29.73 & 0.874 & 64.62 \\
        0.05 & 0.1  & 0.077 & \textbf{0.195} & 27.00 & 0.853 & \underline{30.06} & 0.060 & 0.093 & 29.70 & 0.865 & 62.90 \\
        0.05 & 0.2  & \textbf{0.072} & 0.197 & \underline{27.87} & 0.859 & 32.42 & 0.086 & \textbf{0.083} & 24.37 & 0.876 & 63.28 \\
        \midrule
        0.1  & 0   & 0.077 & 0.210 & 27.15 & 0.861 & 31.27 & 0.061 & 0.111 & 29.27 & 0.873 & 64.71 \\
        0.1  & 0.1  & \underline{0.073} & 0.197 & 27.33 & 0.862 & 31.09 & 0.071 & 0.091 & 26.88 & 0.876 & 63.44 \\
        0.1  & 0.2  & \underline{0.073} & 0.197 & 27.42 & 0.855 & 33.70 & 0.083 & \underline{0.090} & 24.75 & 0.868 & 62.42 \\
        \bottomrule
    \end{tabularx}
    }
\end{table}

\begin{table}[t]
    \caption{Evaluating the application of our post-training framework on cBBDM.}
    \label{tab:cbbdm_performance}
    \centering
    {\fontsize{8}{9.6}\selectfont
    \renewcommand{\arraystretch}{1} 
    \begin{tabularx}{\textwidth}{X >{\centering\arraybackslash}p{0.865cm} >{\centering\arraybackslash}p{0.865cm} >{\centering\arraybackslash}p{0.865cm} >{\centering\arraybackslash}p{0.865cm} >{\centering\arraybackslash}p{0.865cm} >{\centering\arraybackslash}p{0.865cm} >{\centering\arraybackslash}p{0.865cm} >{\centering\arraybackslash}p{0.865cm} >{\centering\arraybackslash}p{0.865cm} >{\centering\arraybackslash}p{0.865cm}}  
        \toprule
        & \multicolumn{5}{c}{\textbf{DWI}} & \multicolumn{5}{c}{\textbf{ADC}} \\
        \cmidrule(lr){2-6} \cmidrule(lr){7-11}
        \textbf{Model} & MAE$\downarrow$ & \makecell{Lesion\\MAE$\downarrow$} & PSNR$\uparrow$ & \makecell{MS-\\SSIM$\uparrow$} & FID$\downarrow$ & MAE$\downarrow$ & \makecell{Lesion\\MAE$\downarrow$} & PSNR$\uparrow$ & \makecell{MS-\\SSIM$\uparrow$} & FID$\downarrow$ \\
        \midrule
        cBBDM & 0.093 & 0.226 & 25.21 & 0.800 & 32.32 & 0.098 & 0.139 & 23.63 & 0.777 & 66.98 \\
        \textbf{cBBDM-PT} & \textbf{0.078} & \textbf{0.221} & \textbf{27.40} & \textbf{0.857} & \textbf{31.55} & \textbf{0.074} & \textbf{0.120} & \textbf{26.62} & \textbf{0.866} & \textbf{66.67} \\
        \bottomrule
    \end{tabularx}
    }
\end{table}

\section{Conclusion}
In this study, we present a novel post-training framework for LDMs in medical images to improve image quality and lesion delineation, generating more realistic and clinically accurate images. By integrating medical image space objectives, our method addresses the challenge of capturing high-frequency details in LDMs. Evaluation on a diffusion MRI-CTP paired dataset of acute ischemic stroke patients demonstrates that our framework surpasses state-of-the-art models in both overall image fidelity and ischemic lesion conspicuity. These results underscore the effectiveness of our framework in enhancing diagnostic reliability and its potential to support clinical decision-making. Moreover, its consistent performance across various LDMs suggests its broad applicability to diverse medical image translation tasks.
\begin{credits}
\subsubsection{\ackname} 
This work was supported by the National Research Foundation of Korea (NRF) grant funded by the Korea government (MSIT) (No. RS-2023-00251022) (K.S.C.); the Phase III (Postdoctoral fellowship) grant of the SPST (SNU-SNUH Physician Scientist Training) Program (K.S.C.); the SNUH Research Fund (No. 04-2024-0600) (K.S.C.); and the Korea Health Technology R\&D Project through the Korea Health Industry Development Institute (KHIDI) grant funded by the Ministry of Health\&Welfare (No. RS-2024-00439549) (K.S.C.).

\subsubsection{\discintname}
The authors have no competing interests to declare that are relevant to the content of this article.
\end{credits}

\bibliographystyle{splncs04}
\bibliography{7_refs}
\end{document}